\documentclass{article}
\pdfoutput=1
\usepackage{spconf,amsmath,graphicx}
\usepackage{amssymb}
\usepackage{mathrsfs}
\usepackage{epstopdf}
\usepackage{algorithmic}
\usepackage{algorithm}
\usepackage{subfigure}
\usepackage{setspace}
\usepackage{url}
\title{Hyperspectral Unmixing with Endmember Variability using semi-supervised Partial Membership Latent Dirichlet Allocation }
\name{Sheng Zou\textsuperscript{$\dagger$}, Hao Sun\textsuperscript{*}, Alina Zare\textsuperscript{$\dagger$}\thanks{The authors wish to thank the National Geospatial-Intelligence Agency for support of this research under the project entitled ``NIP: Functions of Multiple Instances for Hyperspectral Analysis.''}}
\address{Department of Electrical and Computer Engineering, University of Florida\textsuperscript{$\dagger$} \\
Department of Electrical and Computer Engineering, University of Missouri\textsuperscript{*}}
\begin{document}
%\ninept
%
\maketitle
\begin{abstract}
A semi-supervised Partial Membership Latent Dirichlet Allocation approach is developed for hyperspectral unmixing and endmember estimation while accounting for spectral variability and spatial information. Partial Membership Latent Dirichlet Allocation is an effective approach for spectral unmixing while representing spectral variability and leveraging spatial information. In this work, we extend Partial Membership Latent Dirichlet Allocation to incorporate any available (imprecise) label information to help guide unmixing.  Experimental results on two hyperspectral datasets show that the proposed semi-supervised PM-LDA can yield improved hyperspectral unmixing and endmember estimation results. 
\end{abstract}
\begin{keywords}
semi-supervised, partial membership, latent dirichlet allocation, PM-LDA, hyperspectral, unmixing, superpixel, endmember, spectral variability
\end{keywords}
\section{Introduction}
\label{sec:intro}

Hyperspectral unmixing aims to decompose a hyperspectral image cube into constituent endmembers and estimate the proportion of each endmember in each pixel \cite{keshava2002spectral}. A widely used model in hyperspectral unmixing model is the Linear Mixture Model (LMM) that assumes the spectral signature of each pixel is a convex combination of endmember signatures \cite{bioucas2012hyperspectral}. In general, the overwhelming majority of unmixing methods represent endmembers as a single spectral signature. However, in practice, endmembers across a scene exhibit spectral variability \cite{zare2014endmember} due to the effects of varying illumination \cite{bateson1996method}, environmental \cite{asner2000impact, okin2001practical} and atmospheric conditions and the intrinsic variability of materials \cite{zhang2006intra}.   LMM-based algorithms that represent endmembers as a single spectral signature (and do not account for spectral variability) often have reduced performance in proportion estimation \cite{zare2014endmember}.  Thus, a number of methods that perform spectral unmixing while accounting for spectral variability have been developed in the literature \cite{zare2014endmember,somers2011endmember}.

Spectral unmixing methods that account for endmember variability can be roughly paritioned into two major categories: (1) \textit{endmembers as sets} approaches and (2) \textit{endmembers as statistical distributions} approaches. Under ``endmembers as sets,'' the most prominent approach is the multiple endmember spectral mixture analysis (MESMA) \cite{roberts1998mapping, powell2007sub, franke2009hierarchical}, in which each pixel is represented by a collection of endmembers selected from a partitioned spectral library. Each partition corresponds to a set of spectral signatures that represent each endmember and its variations.   For ``endmembers as statistical distributions,'' each endmember is modeled as a random variable according to some statistical distribution.  A commonly chosen endmember distribution is the normal distribution which, when combined with the LMM, results in the Normal Compositional Model (NCM) \cite{stein2003application} for spectral unmixing. The NCM assumes each pixel signature is a convex combination of $K$ endmember variants drawn from $K$ Normal endmember distributions. 
NCM-based methods include Eches \textit{et al}.,\cite{eches2010bayesian, eches2010estimating} and Kazianka \cite{kazianka2012objective} approaches which propose a Markov Chain Mote Carlo (MCMC) sampler to estimate proportion values and endmember covariances under NCM given endmember mean signatures or a spectral library.  Zare \textit{et al}.,\cite{zare2010pce, zare2013sampling} presented MCMC sampler approach to estimate endmember spectral means and proportion values given endmember covariances under an NCM model.  Zhang \textit{et al}.,\cite{zhang2014pso} introduced a particle swarm optimization expectation maximization (PSO-EM) method to estimate the endmember spectral means, endmembers covariances and proportion values. Sheng \textit{et al}., \cite{zou2016hyperspectral} introduced the Partial Membership Latent Dirichlet Allocation (PM-LDA) unmixing approach to estimate all endmember distributions and proportion values under the NCM while leveraging spatial information. 

\begin{figure*}[h]
\begin{center}
\subfigure[]{\includegraphics[height=8cm]{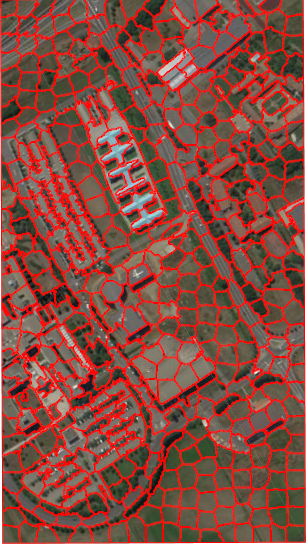}}
\subfigure[]{\includegraphics[height=8cm]{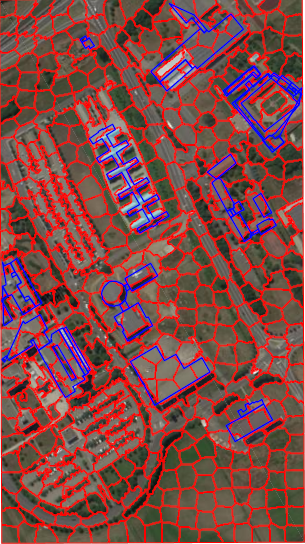}}
\subfigure[]{\includegraphics[height=8cm]{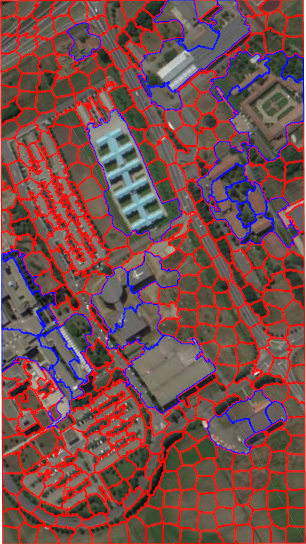}}
\caption{Superpixel segmentation of the Pavia University hyperspectral datacube: (a) superpixel segmentation obtained using SLIC; (b) red: superpixels from SLIC, blue: open street map information; (c) Superpixel segmentation after merging superpixels that intersect the common building or road outlines found in the OpenStreetMap.org map information.}
\label{fig:superpixels}
\end{center}
\end{figure*}

In this paper, we propose the semi-supervised PM-LDA (sPM-LDA) ``endmember as statistical distributions'' approach for spectral unmixing. sPM-LDA is an extension of PM-LDA \cite{zou2016hyperspectral} that provides the ability to leverage any available (even, imprecise) label information to guide unmixing. sPM-LDA also leverages spatial information by encouraging  spatial homogeneity in local regions of the resulting proportion maps. 

%The incorporation of supervision to LDA have been proposed in literature. The goal for introducing supervision is to restrict the topics for each documents. Supervised LDA (cite) limits only one topic for each document. Labeled LDA (cite) is not constrained to one topic per document, it constrains the topic model to use only those topics that correspond to a document's (observed) label set. \par

%Introduction to Open Street Map and SLIC. \par

\section{Motivation}
\label{sec:moti}
% why do we need supervision(less candidate endmembers)
% what's the difference(advantage) of proposed supervison
% why PM-LDA has limitation

In the literature, hyperspectral unmixing methods that are labeled as ``supervised unmixing'' \cite{eches2010bayesian, eches2010estimating}  are generally approaches that rely on prior knowledge of the endmember signatures (\emph{e.g.}, endmembers obtained either from spectral library or an separate endmember extraction algorithm).  In these methods, the hyperspectral unmixing performance is highly dependent on the quality of the spectral library or endmember extraction method used.  In this paper, as opposed to incorporating supervision through prior identification of endmember signatures, sPM-LDA provides an avenue for supervising the proportion values to be estimated.  Since, however, labeling individual pixels with accurate sub-pixel proportion values is, in most cases, infeasible, sPM-LDA leverages \emph{imprecise} proportion labeling.  We propose a ``lazy'' supervision where neither the explicit endmember signatures nor pixel-level proportion values are needed. Instead, sPM-LDA allows for leveraging any available knowledge about which approximate spatial regions share may a common endmember and which do not. More specifically, we segment a hyperspectral image cube into small spatially-contiguous regions using an  superpixel segmentation algorithm as illustrated in Fig. \ref{fig:superpixels}(a). Then, binary labels, $\boldsymbol{\tau}$,  are assigned to a subset of these superpixels where $\boldsymbol{\tau}$ is a $K \times C$ binary matrix (\emph{i.e.}, consisting  only of `1' and `0' values), $K$ is the number of endmembers, and $C$ is the number of superpixels. The value $\tau_{ij}$ is labeled as `1' if endmember $i$ may be found in anywhere within superpixel $j$, otherwise, it is labeled as `0'.  For example, if we want to provide supervision for identifying the `blue roof' endmember in Fig. \ref{fig:superpixels}, we simply label any superpixels that may contain blue roof as `1' in endmember label matrix and `0' otherwise. In practice, these superpixel regions can be labeled using an easy manual labeling method (\emph{i.e.} point-and-click on spatial regions that may contain blue roof) or through an automated approach by identifying the superpixels that intersect the outline of the building in a provided map (\emph{e.g.}, from vector map sources such as OpenStreetMap.org \cite{OSM}).  \par

\section{Hypersepctral Image Superpixel Segmentation}
% background for superpixel segmentationand open streeen map
%%%%%% hyperspectral superpixel review
%Many state-of-the-art image segmentation algorithms have been developed for the gray-scale or RGB imagery. However, relatively little previous work exists for the hyperspectral image superpixel segmentation. In \cite{sethi2015scalable}, the ultrametric contour map (UCM) algorithm was extended to hyperspectral imagery. Brightness, color and texture features were extracted and used to estimate the probability of the existence of a boundary between each pixel in the image. Principal component analysis (PCA) was then used to reduce the image  dimensionality to three dimensions. Then regular UCM superpixel estimation is applied to the PCA reduced data. The normalized cuts algorithm has also been extended for hyperspectral imagery \cite{gillis2012hyperspectral}. This approach combines the spectral and spatial information and uses contour and texture features to partition the given graph recursively while minimizing the cost of the cut at the partition boundaries. In \cite{thompson2010superpixel}, superpixels have been computed with a graph based approach on a grid using the sum of squared differences between neighboring pixels.  
Many state-of-the-art image segmentation algorithms have been developed for the gray-scale or RGB imagery. However, relatively little previous work exists for the hyperspectral image superpixel segmentation. The ultrametric contour map (UCM) algorithm is one of the popular superpixel segmentation algorithms for RGB imagery. Brightness, color and texture features are extracted and used to estimate the probability of the existence of a boundary between each pixel in the image. In \cite{sethi2015scalable}, UCM was extended to hyperspectral imagery by first applying principal component analysis to reduce the image dimensionality to three dimensions then UCM can be directly applied. The normalized cuts algorithm has also been extended for hyperspectral imagery \cite{gillis2012hyperspectral}. This approach combines the spectral and spatial information and uses contour and texture features to partition the given graph recursively while minimizing the cost of the cut at the partition boundaries. In \cite{thompson2010superpixel}, superpixels have been computed with a graph based approach on a grid using the sum of squared differences between neighboring pixels.

In this paper, we consider a hyperspectral extension of Simple Linear Iterative Clustering (SLIC) \cite{achanta2012slic}. The original SLIC algorithm has been widely used for segmenting RGB and gray-scale images. SLIC performs a local clustering of pixels in 5-D space with features as $L$,$a$,$b$ value and the coordinates of the pixel. A $Lab$ color space is a color-opponent space with dimensions $L$ for lightness and $a$ and $b$ for the color-opponent dimensions, based on nonlinearly compressed (e.g. CIE XYZ) coordinates \cite{connolly1997study}.

\section{Partial Membership Latent Dirichlet Allocation}
\label{sec:pmlda}

% 1/2 -3/4 pages
% LDA (equations)\\
% PM-LDA (equations)

Partial Membership Latent Dirichlet Allocation (PM-LDA) \cite{chen2016partial,chenpartial} is an extension of Latent Dirichlet Allocation topic modeling \cite{blei2003latent} that allows \emph{words} to have partial membership across multiple topics.  The use of partial memberships allows for topic modeling given data sets in which crisp topic assignments (as done by LDA) is insufficient since data points (or words) may straddle multiple topics simultaneously.  

\begin{figure}[h]
\begin{center}
\includegraphics[height=4cm]{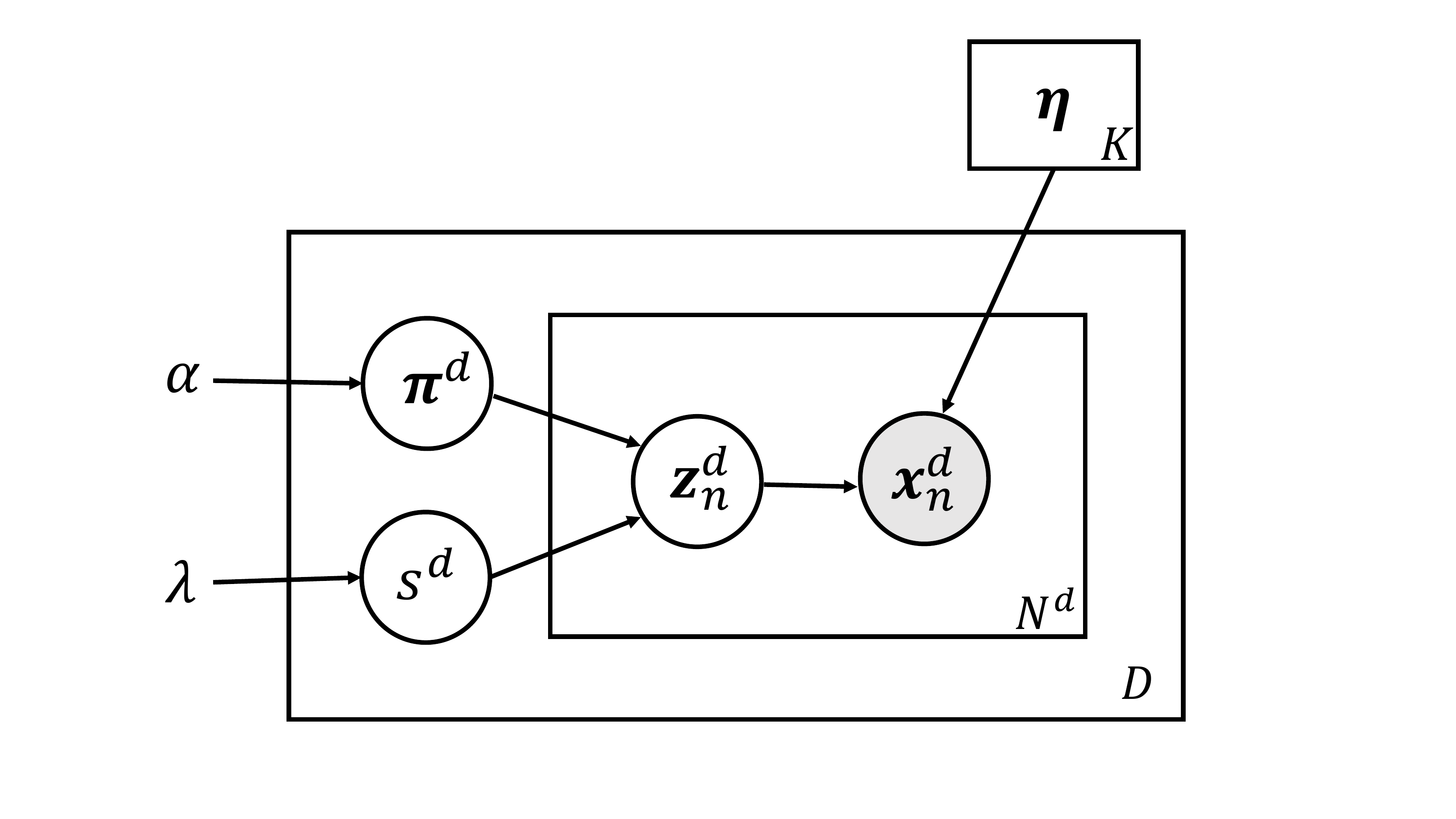}
\caption{Graphical model of PM-LDA}
\label{fig:pmlda}
\end{center}
\end{figure}

 PM-LDA  is a hierarchical Bayesian model in which data in a \emph{corpus} is organized at two levels: the \emph{word} and \emph{document} level, as illustrated in Fig. \ref{fig:pmlda}.
 In the PM-LDA model, the random variable associated with a data point, $\mathbf{x}$, is assumed to be distributed according to multiple topics with a continuous partial membership in each topic, $\mathbf{z}$. Specifically, the PM-LDA model is
\begin{eqnarray}
p(\boldsymbol{\pi}^d,s^d, \mathbf{z}^d_n,\mathbf{x}^d_n|\boldsymbol{\alpha},\lambda,\boldsymbol{\beta})&=&p(\boldsymbol{\pi}^d|\boldsymbol{\alpha})p(s^d|\lambda)p(\mathbf{z}^d_n|\boldsymbol{\pi}^d,s^d)\nonumber \\
&& \prod_{k=1}^{K}p_k(\mathbf{x}_n^d|\eta_k)^{{z}^d_{nk}} 
%eta
\label{eqn:pmlda}
\end{eqnarray} 
%where $K$ is the number of topics in corpus, $\beta_k$ is the parameter set for word distribution of $k$th topic. $\eta_k$ is the natural parameter converted from $\beta_k$. $\mathbf{x}^d_n$ is the $n$th word in document $d$, $\mathbf{x}^d_n  \sim \text{Expon}(\sum_{k} {z}^d_{nk} \eta_k)$, $\mathbf{z}^d_n\sim \text{Dir}(\boldsymbol{\boldsymbol{\pi}^ds^d}) $ is the partial membership vector of $\mathbf{x}^d_n$, ${z}^d_{nk}$ is the $k$th entry of  $\mathbf{z}^d_n$,  $\boldsymbol{\pi}^d \sim \text{Dir}(\boldsymbol{\alpha})$ and $s^d \sim \text{exp}(\lambda)$ are the topic proportion vector and  the level of topic mixing in document $d$, respectively. 
%
where $K$ is the number of topics in a corpus,  $\boldsymbol{\pi}^d \sim \text{Dir}(\boldsymbol{\alpha})$ is the topic proportion vector that provides the proportion of each topic found in document $d$, and $s^d \sim \text{exp}(\lambda)$ controls the variance around the topic proportion mean.  The variable $\mathbf{x}^d_n$ is the $n$th word in document $d$  and is distributed according to $\prod_{k=1}^{K}p_k(\mathbf{x}_n^d|\eta_k)^{{z}^d_{nk}}$ where  $p_k(\cdot | \eta_k)$ is the distribution governing the $k$th topic,  $\eta_k$ are the associated topic distribution parameters, $\mathbf{z}^d_n\sim \text{Dir}(\boldsymbol{\boldsymbol{\pi}^ds^d}) $ is the partial membership vector of $\mathbf{x}^d_n$, and ${z}^d_{nk}$ is the $k$th entry of  $\mathbf{z}^d_n$.  If $p_k(\cdot | \eta_k)$ is assumed to be an exponential family distribution, then  $\mathbf{x}^d_n  \sim \text{Expon}(\sum_{k} {z}^d_{nk} \eta_k)$ as shown in \cite{heller2008statistical}. 

Given the data set (which has been partitioned into documents), $\mathbf{X} = \left\{ \mathbf{X}^1, \mathbf{X}^2, \ldots, \mathbf{X}^D\right\}$, and hyperparameters $\mathbf{\Psi} = \{\mathbf{\alpha}, \lambda\}$, the goal of parameter estimation given the PM-LDA model is to estimate the topic proportion of each document, $\boldsymbol{\pi}^d$, the topic mixing level in each document, $s^d$, the partial memberships of each word in each topic, $\mathbf{z}^d_n$, and the natural parameters defining the probability distribution of each topic, $\eta_k$.
\begin{algorithm}[ht]
\begin{algorithmic}[1]
\REQUIRE{A corpus $\mathbf{D}$, the number of topics $K$, hyperparameters $\mathbf{\Psi} = \{\mathbf{\alpha}, \lambda\}$,  and the number of iterations $T$}
\ENSURE{Collection of all samples: $\boldsymbol{\Pi}^{(t)}, \mathbf{S}^{(t)}, \mathbf{M}^{(t)}$,  $\boldsymbol{\beta}^{(t)}$  
\FOR{$t=1:T$}
\FOR{$d=1:D$}
\STATE \underline{Sample $\boldsymbol{\pi}^d$:} Draw candidate: $\boldsymbol{\pi}^\dagger \sim \text{Dir}(\boldsymbol{\alpha})$ \\
 Accept candidate with probability:\\ $a_{\boldsymbol{\pi}}=\min \left \{ 1, \frac{p(\boldsymbol{\pi}^{\dagger}, s^{(t-1)}, \mathbf{Z}^{(t-1)}, \mathbf{X}|\boldsymbol{\Psi}) p(\boldsymbol{\pi}^{(t-1)}|\boldsymbol{\alpha})}{ p(\boldsymbol{\pi}^{(t-1)}, s^{(t-1)}, \mathbf{Z}^{(t-1)}, \mathbf{X}|\boldsymbol{\Psi}) p(\boldsymbol{\pi}^\dagger|\boldsymbol{\alpha})}\right \}$

\STATE \underline{Sample $s^d$:} Draw candidate: $s^\dagger \sim \text{exp}(\lambda)$\\
Accept candidate with probability:\\
$a_s=\min \left \{ 1, \frac{p(\boldsymbol{\pi}^{(t)}, s^{\dagger}, \mathbf{Z}^{(t-1)}, \mathbf{X}|\boldsymbol{\Psi})p(s^{(t-1)}|\lambda)}{p(\boldsymbol{\pi}^{(t)}, s^{(t-1)}, \mathbf{Z}^{(t-1)}, \mathbf{X}|\boldsymbol{\Psi})p(s^{\dagger}|\lambda)}\right \}$

\FOR{$n=1:N^d$}
\STATE \underline{Sample $\mathbf{z}^d_n$:} Draw candidate: $\mathbf{z}_n^\dagger \sim \text{Dir}(\mathbf{1}_K)$\\
Accept candidate with probability:\\
$a_{\mathbf{z}}=\min \left\{1, \frac{p(\boldsymbol{\pi}^{(t)}, s^{(t)}, \mathbf{z}_n^\dagger, \mathbf{x}_n|\boldsymbol{\Psi})}{p(\boldsymbol{\pi}^{(t)}, s^{(t)}, \mathbf{z}_n^{(t-1)}, \mathbf{x}_n|\boldsymbol{\Psi})}\right\} $
\ENDFOR
\ENDFOR

\FOR{$k=1:K$}
\STATE \underline{Sample $\mu_k$:} Draw proposal: ${\mu}_k^{\dagger}\sim\mathcal{N}(\cdot|{\mu}_{\mathbf{D}},{\Sigma}_{\mathbf{D}})$\\ 
${\mu}_{\mathbf{D}}$ and ${\Sigma}_{\mathbf{D}}$ are  mean and covariance of the data\\
Accept candidate with probability:\\
$a_k=\min\left\{1, \frac{p\left(\boldsymbol{\Pi}^{(t)}, \mathbf{S}^{(t)}, \mathbf{M}^{(t)}, \mathbf{D}|{\mu}_k^{\dagger}\right)\mathcal{N}(\mu_k^{(t-1)}|\mu_\mathbf{D}, \Sigma_\mathbf{D})}{p\left(\boldsymbol{\Pi}^{(t)}, \mathbf{S}^{(t)}, \mathbf{M}^{(t)}, \mathbf{D}|\boldsymbol{\mu}_k^{(t-1)}\right)\mathcal{N}(\mu_k^{\dagger}|\mu_\mathbf{D}, \Sigma_\mathbf{D})} \right\}$
\ENDFOR
\STATE \underline{Sample covariance matrices $\boldsymbol{\Sigma}= \sigma^2\mathbf{I}$:}\\
Draw candidate from: $\sigma^2 \sim$ Unif$(0,u)$\\ with $u = \frac{1}{2}\left\{\max_{\mathbf{x}_n}d^2(\mathbf{x}_n-{\mu}_{\mathbf{D}})-\min_{\mathbf{x}_n}d^2(\mathbf{x}_n-{\mu}_{\mathbf{D}}) \right\}$\\
Accept candidate with probability: \\
$a_{\boldsymbol{\Sigma}}= \min\left\{1, \frac{p\left(\boldsymbol{\Pi}^{(t)}, \mathbf{S}^{(t)}, \mathbf{M}^{(t)}, \mathbf{D}|\boldsymbol{\Sigma}^\dagger\right)}{p\left(\boldsymbol{\Pi}^{(t)}, \mathbf{S}^{(t)}, \mathbf{M}^{(t)}, \mathbf{D}|\boldsymbol{\Sigma}^{(t-1)}\right)} \right\}.$
\ENDFOR}
\end{algorithmic}
\caption{Metropolis-within-Gibbs Sampling Method for Parameter Estimation}
\label{alg:pmlda}
\end{algorithm}

PM-LDA has been used previously for NCM-based hyperspectral unmixing and endmember estimation \cite{zou2016hyperspectral}.  PM-LDA is applied to unmixing by, first, over-segmenting the hyperspectral scene into spatially-contiguous superpixels, as shown in Fig. \ref{fig:superpixels}(a).  Each superpixel is assumed to be a \emph{document} in the PM-LDA model.  Given the superpixel segmentation and the assumption that the topic (\emph{i.e.}, endmember) distributions are Gaussian (to assume the NCM), then the parameters of the PM-LDA model can be directly related to parameters of interest in the NCM unmixing model.  Namely, the $K$ topic distributions governed by parameters $\beta_k = \left\{\mu_k, \Sigma_k\right\}$ correspond to the $K$ Gaussian endmember distributions. The cluster parameter $\beta_k = \left\{\mu_k, \Sigma_k\right\}$ can be mapped to the natural parameters of the exponential family distribution,  $\eta_k = \left\{\Sigma_k^{-1}, \Sigma_k^{-1} \mu_k\right\}$. The partial membership vector for data point $n$ in document $d$, $\mathbf{z}_n^d$, is the proportion vector associated with the $n^{th}$ data point in the $d^{th}$ superpixel.  The topic proportion vectors for a document, $\boldsymbol{\pi}^d$, correspond to the average proportion vector for a superpixel with the mixing level $s^d$ corresponding to how much each proportion vector in the document is likely to vary from the average proportion vector.  Thus, an entire hyperspectral scene or collection of hyperspectral scenes can be modeled as a corpus in PM-LDA. 

Alg \ref{alg:pmlda} summarizes a Metropolis-within-Gibbs sampler to perform parameter estimation for PM-LDA where $\boldsymbol{\Pi}^{(t)}$ are the collection of all document-level topic proportions at time $t$,  $\mathbf{S}^{(t)}$ are the topic mixing levels, $\mathbf{M}^{(t)}$ are the collection of partial membership vectors, and  $\boldsymbol{\beta}^{(t)}$ are the parameters defining each topic distribution. A Matlab implementation of PM-LDA parameter estimation can be found on our Github page \cite{pmldacode}.

An advantage of the use of a superpixel segmentation to define documents during unmixing is that it allows us to leverage the expected similarity of the materials found in neighboring pixels.  In other words, spectrally homogeneous neighborhoods are likely to be grouped within a superpixel. Using PM-LDA, all of the pixels in a superpixel are paired with proportion vectors drawn from the same Dirichlet distribution with a shared average proportion vector ($\boldsymbol{\pi}^d$) and the variance around that mean is governed by $s^d$.  Larger $s^d$ values correspond to more spatially and spectrally homogeneous superpixels.

%\section{Supervised LDA}
% background for labeled LDA and the difference compared with proposed method

\section{Proposed Method}
%(1)Supervised PM-LDA goes here.\par 
The proposed semi-supervised partial membership Latent Dirichlet Allocation (sPM-LDA) approach is comprised of three separate steps: 1) Map-Guided Hyperspectral SLIC; 2) Label assignment for each superpixel; and 3) Incorporation of endmember labels into PM-LDA for inference and parameter estimation. \par
%(2)The combination of SLIC and open street map goes here (Hao).\par

\subsection{Step 1: Map-Guided Hyperspectral SLIC}
In step one of the proposed approach, an input hyperspectral image is over-segmentated into superpixel regions using a map-guided segmentation approach.  This superpixel segmentation is used to define spatially-contiguous regions with approximately similar proportion vectors. Furthermore, the map-guidance is used to merge superpixels that are identified to contain the same map objects as well as to autonomously generate map-derived labels for these superpixels.  

To perform the superpixel segmentation, we extend the Simple Linear Iterative Clustering (SLIC) to hyperspectral imagery and to leverage map information. Previously in the literature, dimensionality reduction was applied to HSI prior to allow for application of SLIC \cite{zhang2015slic, sun2017map}. In this paper, no dimensionality reduction is applied.  Instead, the full spectral information is used  in combination with spatial information. The proposed hyperspectral SLIC is iterative: (1) initial cluster centers are obtained with a regular spatial sampling of the image where the cluster centers are vectors in which spectral information is concatenated with spatial coordinates; (2) then, each pixel is assigned to the nearest cluster center using a distance measure that includes a spectral and a spatial term; and (3) cluster centers are then updated to be equal to the average of all pixels assigned to the cluster. This process is iterated until the convergence.

To incorporate the map-guidance, post-processing based on map information (specifically, OpenStreetMap.org (OSM) \cite{haklay2008openstreetmap}) is performed. OpenStreetMap is a collaborative project to create a free editable map of the world. From OSM data, we can parse out linear features and polygons that provide roadway locations and building profiles.  Each of the OSM building profiles and roadway maps are paired with latitude and longitude information that can be used to align the map with the aerial hyperspectral data cube. However, even with world coordinates for the map information, mis-alginment with the hyperspectral data cube is likely, particularly, if the hyperspectral data cube does not have pixel-level accurate geo-location information.  Thus, as opposed to attempting to label or link individual image pixels based on map information, superpixels are merged and labeled instead.  The size of each superpixel provides tolerance for some level of mis-alignment between map data and imagery.   The full map-guided hyperspectral SLIC is summarized in Algorithm \ref{alg:SLIC}.

\begin{algorithm}[ht]
\begin{algorithmic}[1]
    \REQUIRE HSI Data, $K$ (number of superpixels), $m$ (scaling factor)
    % \ENSURE superpixels
\STATE Initialize cluster centers $\left\{\mathbf{c}_k\right\}_{k=1}^K$ by sampling pixels at regular grid steps.
\STATE Perturb  $\left\{\mathbf{c}_k\right\}_{k=1}^K$ in an $n \times n$ neighborhood to the lowest gradient position.
\STATE \textbf{Repeat}
\FOR {each  $\mathbf{c}_k$}
\STATE Calculate the spectral distance of pixel $\mathbf{x}_i$ and $\mathbf{c}_k$ over all bands $\lambda$: $d_{spectral} = \sum^B_{\lambda=1} \| \mathbf{x}_i(\lambda) -  \mathbf{c}_k(\lambda)\|^2_2$
\STATE Calculate the spatial distance of pixel $\mathbf{x}_i$ and $\mathbf{c}_k$:\\ $d_{spatial} = \sqrt{(a_{\mathbf{x}_i}-a_{\mathbf{c}_k})^2+(b_{\mathbf{x}_i}-b_{\mathbf{c}_k})^2}$ where $a,b$ are pixel coordinates.
\STATE   Assign each pixel to a $\mathbf{c}_k$ in a $2S \times 2S$ square neighborhood based on the minimum spectral and spatial distance: $d_{\mathbf{x}_i,\mathbf{c}_k} = d_{spectral} + \frac{m}{S} d_{spatial}$
\ENDFOR
\STATE Update cluster centers as the mean of all pixels assigned to the cluster.
\STATE \textbf{ Until} stopping criterion is reached.
\STATE Align map data to imagery
\FOR {Each Map Polygon}
\STATE Merge all superpixels that overlap with this polygon
\ENDFOR
\end{algorithmic}
\caption{Map-Guided Hyperspectral SLIC Superpixel Segmentation}
\label{alg:SLIC}
\end{algorithm}

%%%%%%%%%%%%%%%%%%%%%%%%%%%%

\subsection{Step 2: Superpixel Labeling}\par
In step two, given the superpixel segmentation, supervision is imposed by assigning an endmember label vector $\tau_{j}$ to each superpixel $j$.  The labels are used to identify the superpixels that \emph{do not} contain a specific endmember.   In other words, if $\tau_{ij} = 1$, then superpixel $j$ may or may not contain endmember $i$.  In contrast, if $\tau_{ij} = 0$, then superpixel $j$ is constrained to not contain endmember $i$ (\emph{i.e.}, the proportion values for endmember $i$ is constrained to be zero for all pixels in the superpixel). This type of labeling is similar in concept to labels using with Multiple Instance Learning approaches \cite{maron1998framework, jiao2015functions}. Given this labeling framework, an endmember material $i$ whose locations are approximately known and can be supervised by labeling each superpixel $j$ that may contain that endmember with $\tau_{ij} = 1$ and all other superpixel pixels with $\tau_{ij}=0$, for example, consider the \emph{red roof} material in the Pavia Univeristy dataset which can be found in several superpixels.  It is relatively easy for a user to label the superpixels containing red roof.

On the contrary, there may be many endmembers whose locations throughout the scene are unknown.  These endmembers are treated as unsupervised. For instance, there are a large number of superpixels in the Pavia University data set containing soil, asphalt, shadow, and vegetation materials. It can be time consuming and challenging to label these accurately. Thus, all superpixels are labeled with `1' for those endmembers as they all have the possibility of containing those endmembers. Considering the extreme case where the endmember label matrix $\tau_{ij} = \mathbf{1_{KC}}$, then the proposed  semi-supervised PM-LDA algorithm degrades to the standard PM-LDA unmixing algorithm. \par

\subsection{Step 3: Using labels in PM-LDA for inference and parameter estimation}

Under the unsupervised PM-LDA algorithm, topic proportions for each superpixel are drawn from a Dirichlet distribution with hyperparameters $\alpha$. To impose supervision, we propose to draw topic proportions from a Dirichlet distribution whose mean is the point-wise product of original Dirichlet distribution mean and the endmember label vector for that superpixel. By applying the binary endmember labels to the topic proportion means,  candidate topic proportion vectors are restricted to have zero entries on the endmembers that are not located in a particular superpixel.  To be more specific, the topic proportion sampling step (line 3, Alg \ref{alg:pmlda}) is changed to $\boldsymbol{\pi}^\dagger \sim \text{Dir}(\tau^{d} \otimes\boldsymbol{\alpha})$, where $\tau^{d}$ denotes the binary vectors of endmember labels for $d$th superpixel and $\otimes$ indicates the point-wise product.  Other than this change, sPM-LDA follows the PM-LDA algorithm (Alg. 1) exactly.

\section{EXPERIMENTS}
\label{sec:experiments}

Two  hyperspectral data sets, \textit{University of Pavia} and \textit{MUUFL Gulfport}, were unmixed using the proposed method. We compared the proposed sPM-LDA with two other NCM-based algorithms (unsupervised) PM-LDA \cite{zou2016hyperspectral} and NCM-Bayes \cite{eches2010bayesian}. 

\subsection{\textit{University of Pavia}}

\begin{figure}[ht]
\begin{center}
\includegraphics[height=8cm]{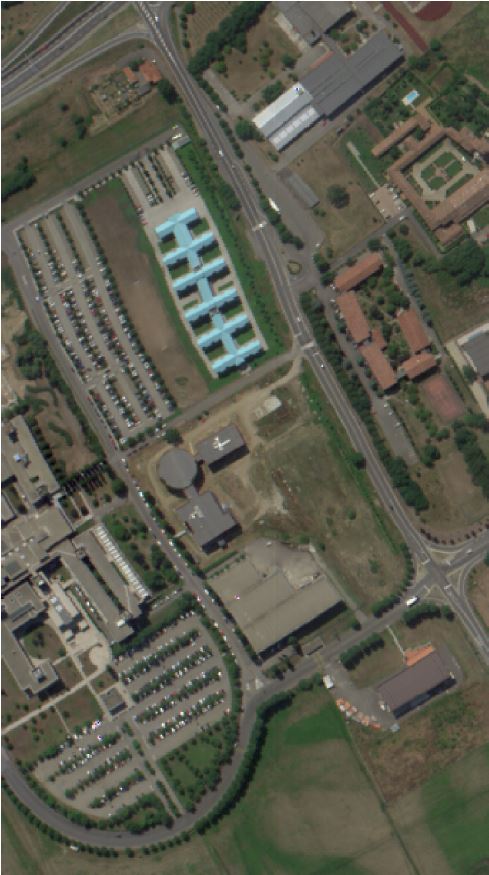}
\caption{RGB-image of the University of Pavia hyperspectral data set}
\label{fig:pavia}
\end{center}
\end{figure}

\begin{figure*}[htb!]
\begin{center}
\subfigure[]{\includegraphics[height=3.5cm]{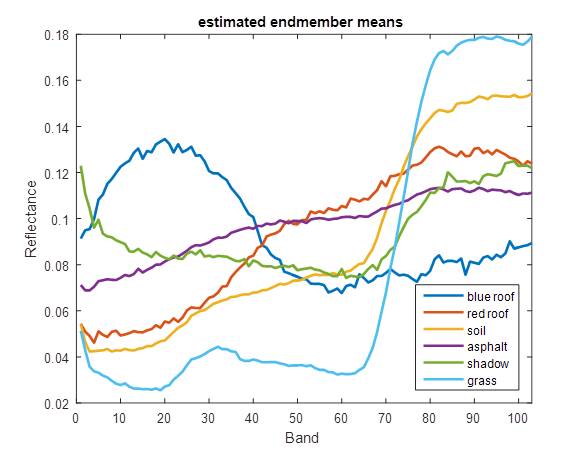}}
\subfigure[]{\includegraphics[height=3.5cm]{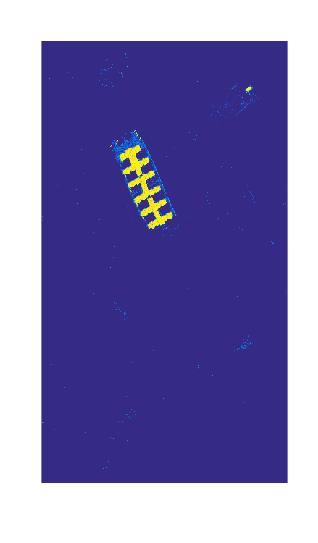}}
\subfigure[]{\includegraphics[height=3.5cm]{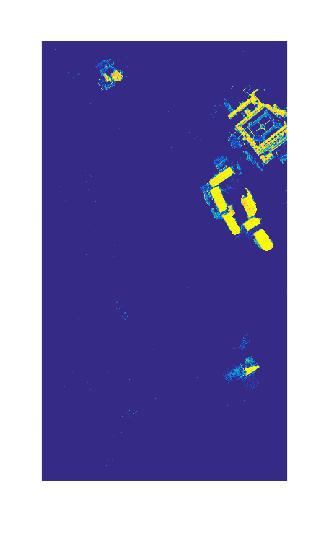}}
\subfigure[]{\includegraphics[height=3.5cm]{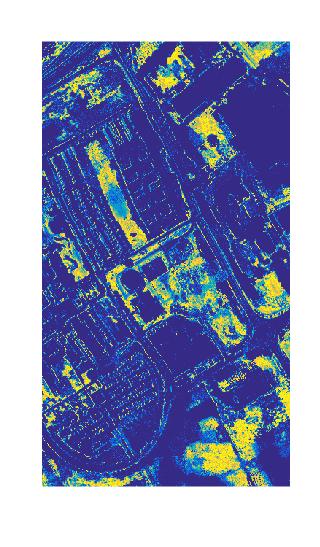}}
\subfigure[]{\includegraphics[height=3.5cm]{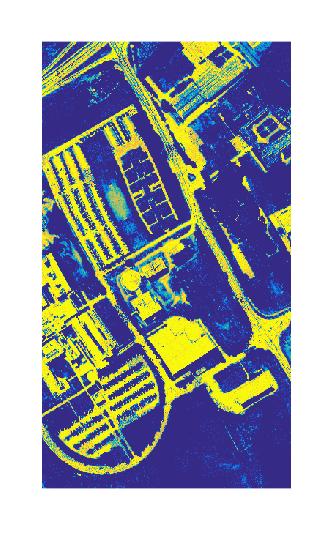}}
\subfigure[]{\includegraphics[height=3.5cm]{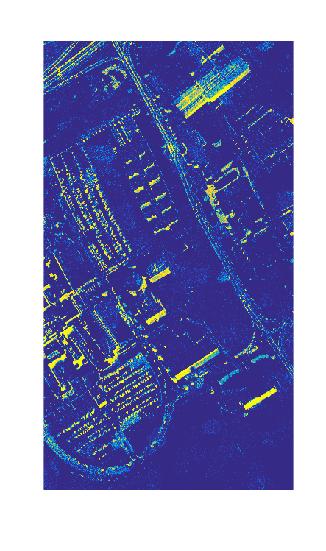}}
\subfigure[]{\includegraphics[height=3.5cm]{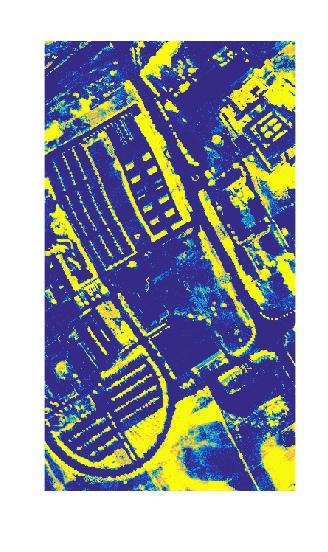}}

\subfigure[]{\includegraphics[height=3.5cm]{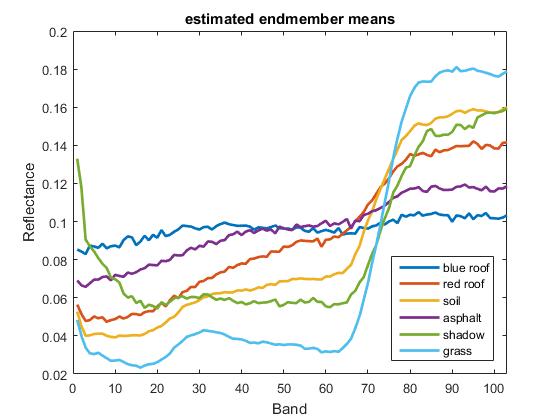}}
\subfigure[]{\includegraphics[height=3.5cm]{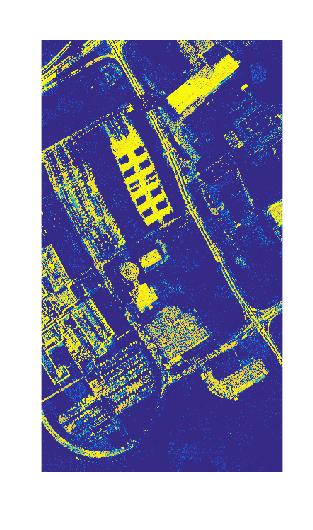}}
\subfigure[]{\includegraphics[height=3.5cm]{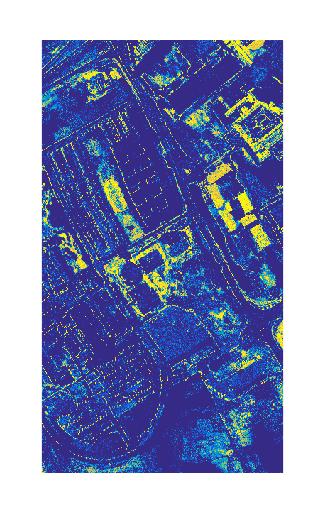}}
\subfigure[]{\includegraphics[height=3.5cm]{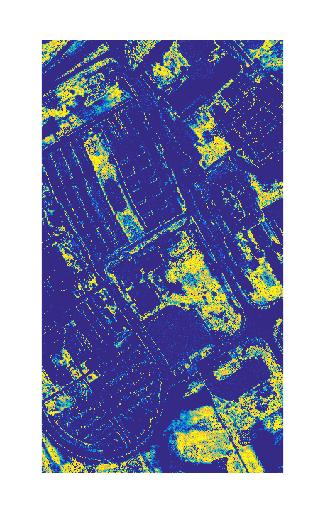}}
\subfigure[]{\includegraphics[height=3.5cm]{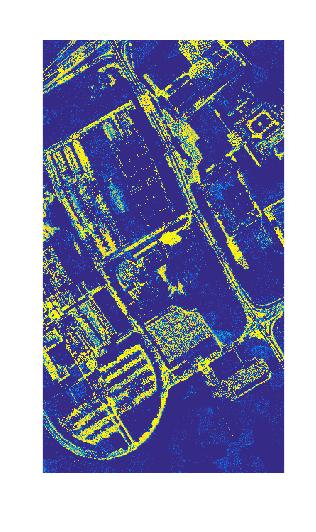}}
\subfigure[]{\includegraphics[height=3.5cm]{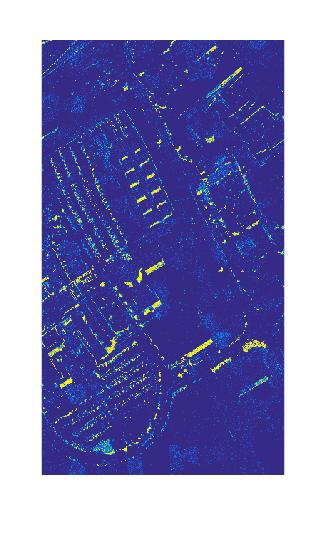}}
\subfigure[]{\includegraphics[height=3.5cm]{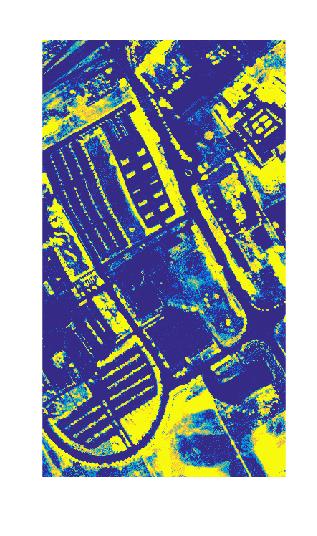}}

\subfigure[]{\includegraphics[height=3.5cm]{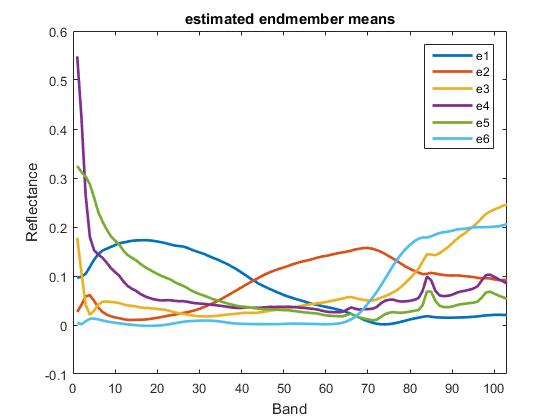}}
\subfigure[]{\includegraphics[height=3.5cm]{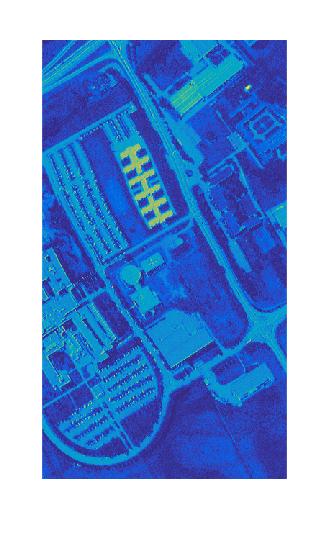}}
\subfigure[]{\includegraphics[height=3.5cm]{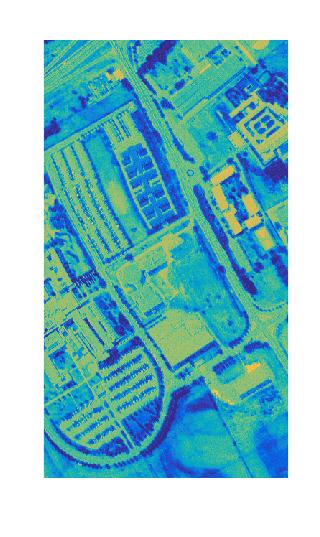}}
\subfigure[]{\includegraphics[height=3.5cm]{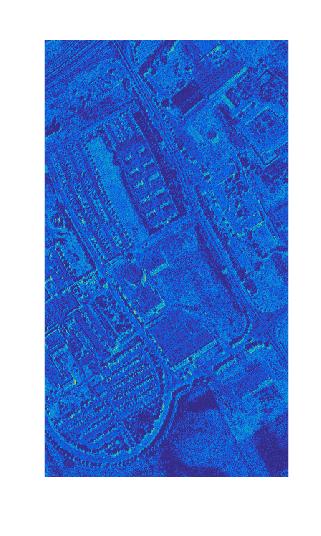}}
\subfigure[]{\includegraphics[height=3.5cm]{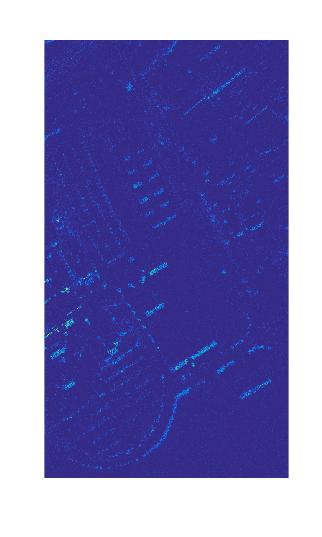}}
\subfigure[]{\includegraphics[height=3.5cm]{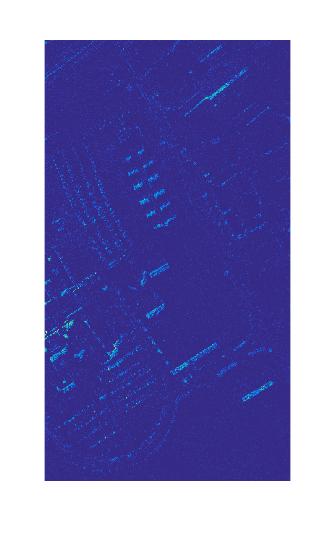}}
\subfigure[]{\includegraphics[height=3.5cm]{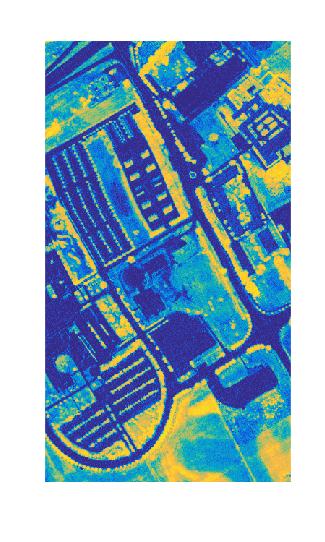}}

\caption[test]{Endmember means estimated by (a) supervised PM-LDA, (h) PM-LDA, (o) VCA. Estimated proportion maps using (b)-(g) supervised PM-LDA, (i)-(n) PM-LDA, (p)-(u) NCM-Bayes.}
\label{fig:p_pavia}
\end{center}
\end{figure*}

\begin{figure*}%[h]
\begin{center}
\subfigure[]{\includegraphics[height=3.5cm]{E.jpg}}
\subfigure[blue roof]{\includegraphics[height=3.5cm]{P1.jpg}}
\subfigure[red roof]{\includegraphics[height=3.5cm]{P2.jpg}}
\subfigure[soil]{\includegraphics[height=3.5cm]{P3.jpg}}
\subfigure[asphalt]{\includegraphics[height=3.5cm]{P4.jpg}}
\subfigure[shadow]{\includegraphics[height=3.5cm]{P5.jpg}}
\subfigure[grass]{\includegraphics[height=3.5cm]{P6.jpg}}

\subfigure[]{\includegraphics[height=3.5cm]{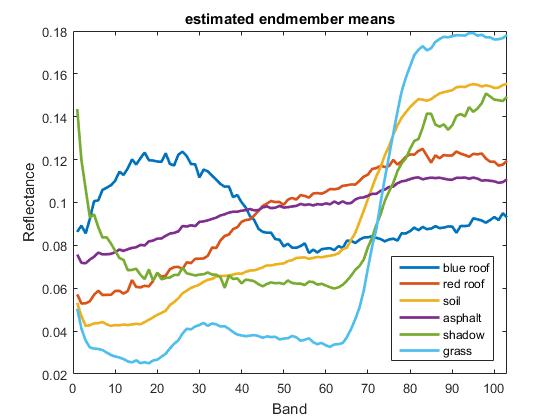}}
\subfigure[blue roof]{\includegraphics[height=3.5cm]{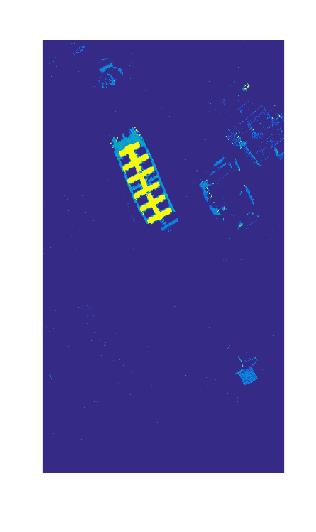}}
\subfigure[red roof]{\includegraphics[height=3.5cm]{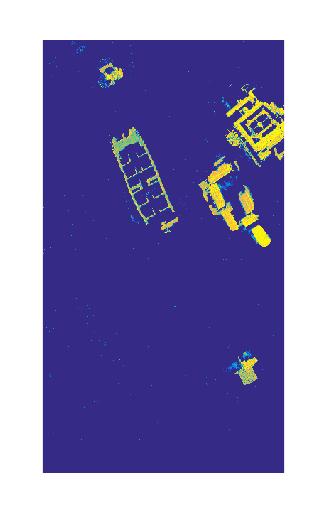}}
\subfigure[soil]{\includegraphics[height=3.5cm]{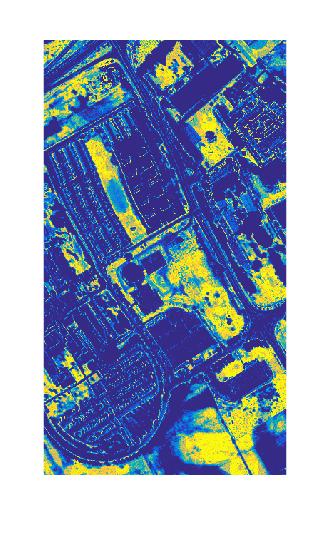}}
\subfigure[asphalt]{\includegraphics[height=3.5cm]{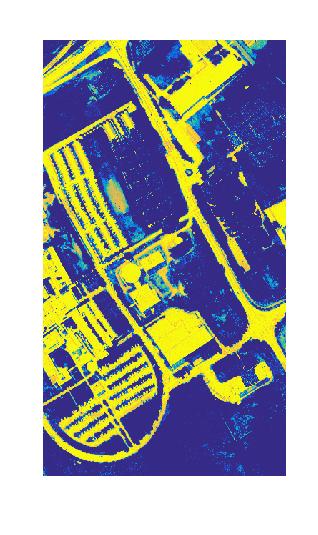}}
\subfigure[shadow]{\includegraphics[height=3.5cm]{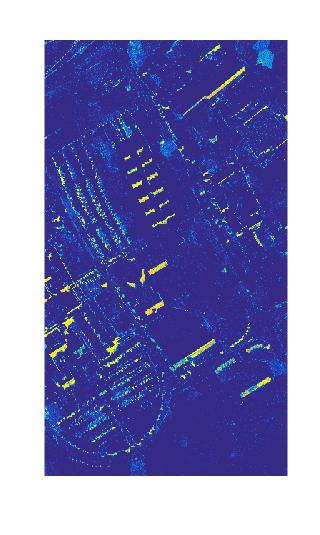}}
\subfigure[grass]{\includegraphics[height=3.5cm]{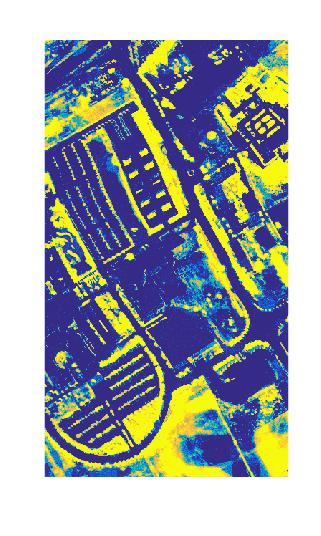}}

\caption[Estimated endmember means and proportion maps on Pavia with or without precise labels]{Endmember means estimated on Pavia by (a) semi-supervised PM-LDA with precise labels, (h) semi-supervised PM-LDA with imprecise labels. Estimated proportion maps on Pavia using (b)-(g) semi-supervised PM-LDA with precise labels, (i)-(n) semi-supervised PM-LDA with imprecise labels.}
\label{fig:p_pavia_labels}
\end{center}
\end{figure*}

\begin{figure*}[ht]
\begin{center}
\subfigure[]{\includegraphics[height=4cm]{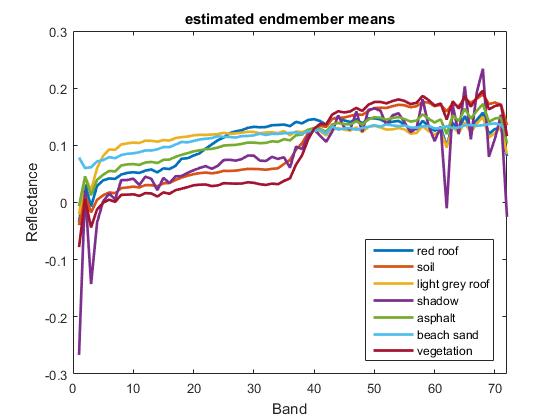}}
\subfigure[]{\includegraphics[height=4cm]{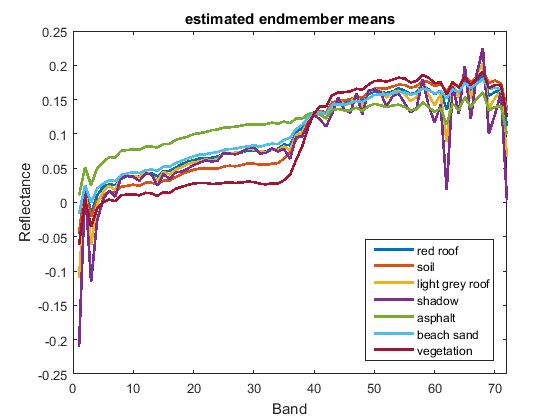}}
\subfigure[]{\includegraphics[height=4cm]{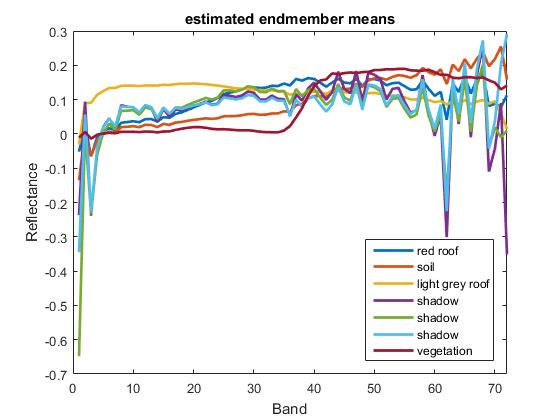}}
\caption{Endmember means estimated by (a) supervised PM-LDA, (b) PM-LDA, (c) VCA.}
\label{fig:e_gulfport}
\end{center}
\end{figure*}

\begin{figure*}[ht]
\begin{center}
\subfigure[]{\includegraphics[height=2cm]{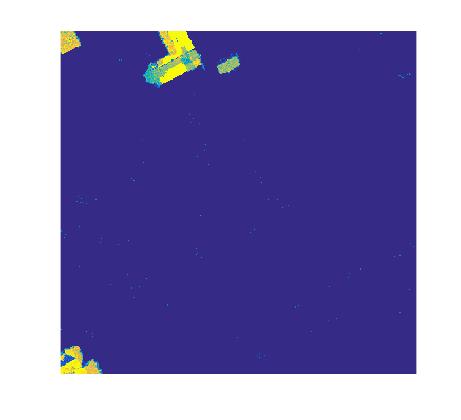}}
\subfigure[]{\includegraphics[height=2cm]{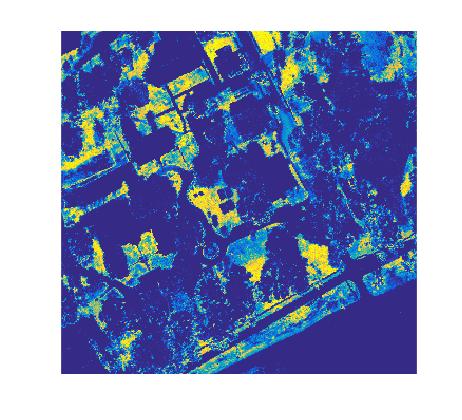}}
\subfigure[]{\includegraphics[height=2cm]{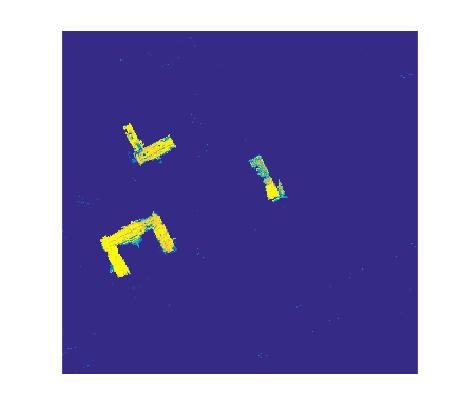}}
\subfigure[]{\includegraphics[height=2cm]{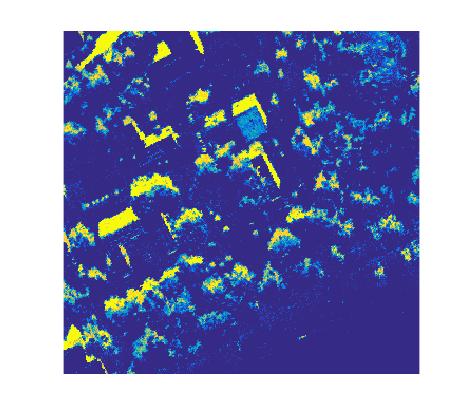}}
\subfigure[]{\includegraphics[height=2cm]{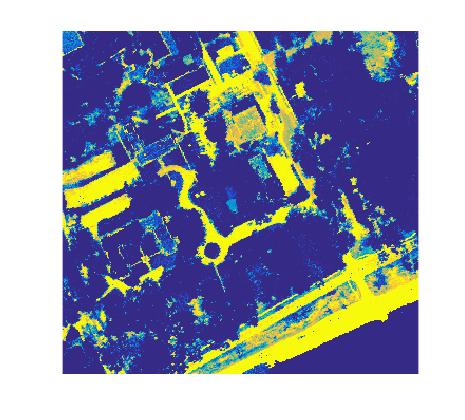}}
\subfigure[]{\includegraphics[height=2cm]{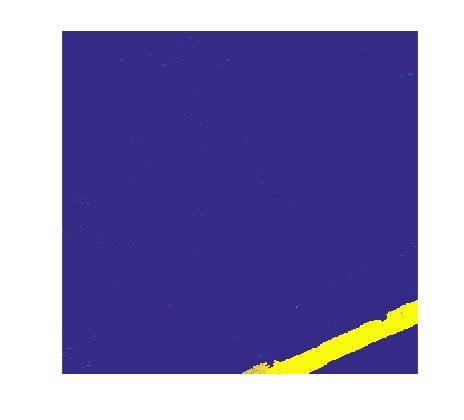}}
\subfigure[]{\includegraphics[height=2cm]{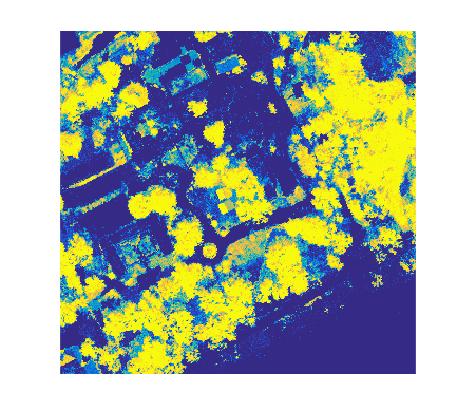}}

\subfigure[]{\includegraphics[height=2cm]{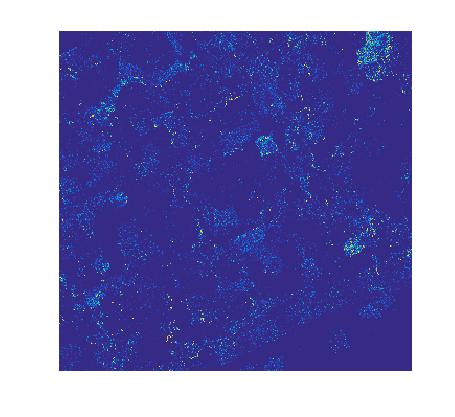}}
\subfigure[]{\includegraphics[height=2cm]{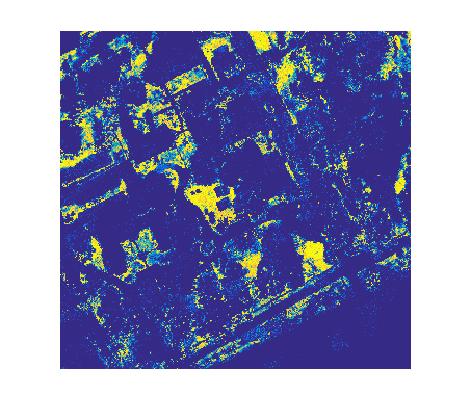}}
\subfigure[]{\includegraphics[height=2cm]{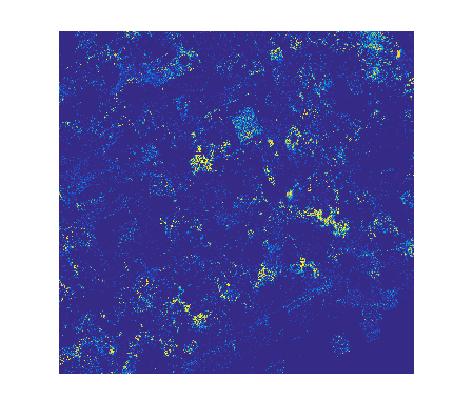}}
\subfigure[]{\includegraphics[height=2cm]{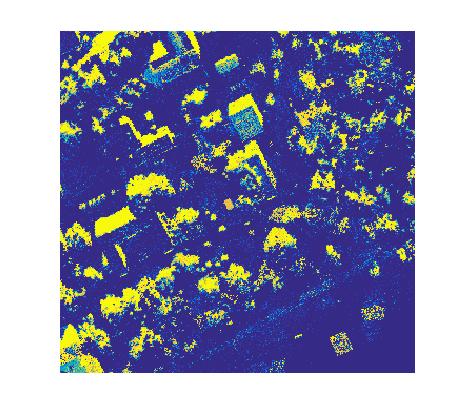}}
\subfigure[]{\includegraphics[height=2cm]{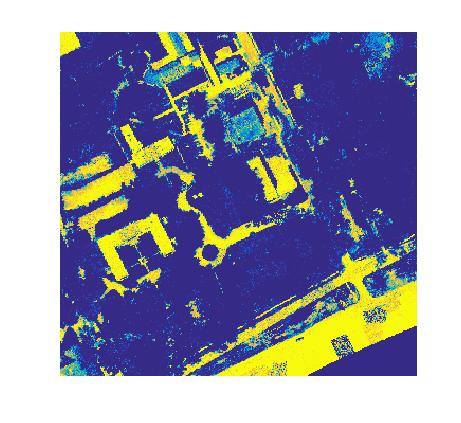}}
\subfigure[]{\includegraphics[height=2cm]{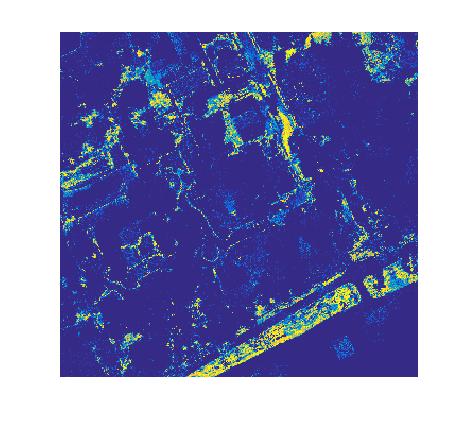}}
\subfigure[]{\includegraphics[height=2cm]{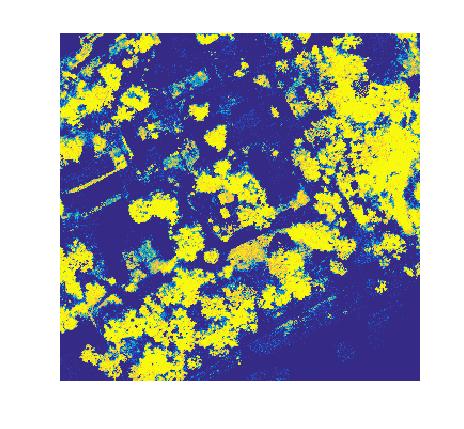}}

\subfigure[]{\includegraphics[height=2cm]{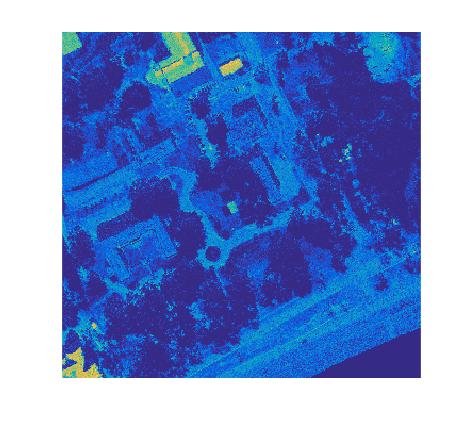}}
\subfigure[]{\includegraphics[height=2cm]{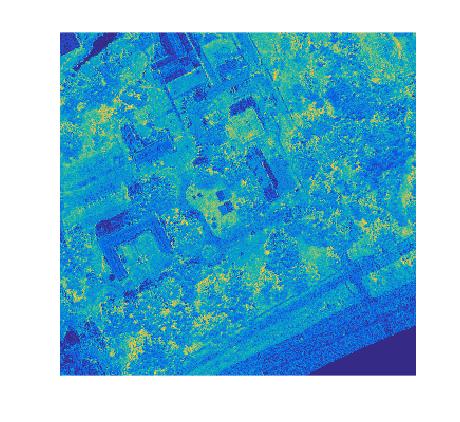}}
\subfigure[]{\includegraphics[height=2cm]{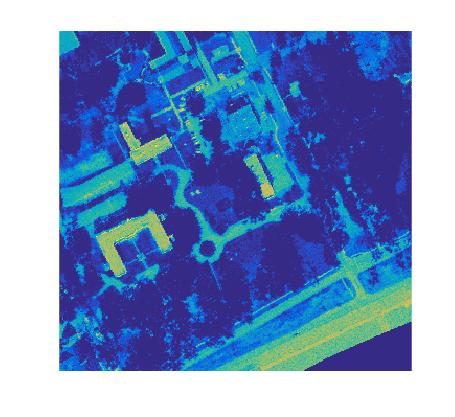}}
\subfigure[]{\includegraphics[height=2cm]{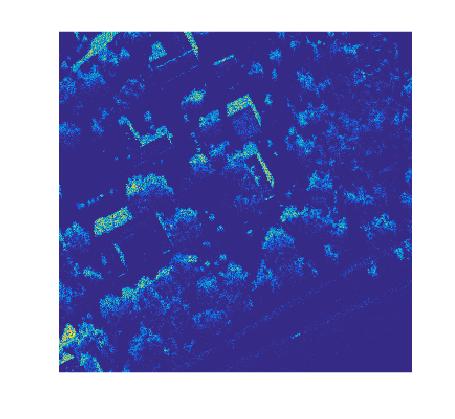}}
\subfigure[]{\includegraphics[height=2cm]{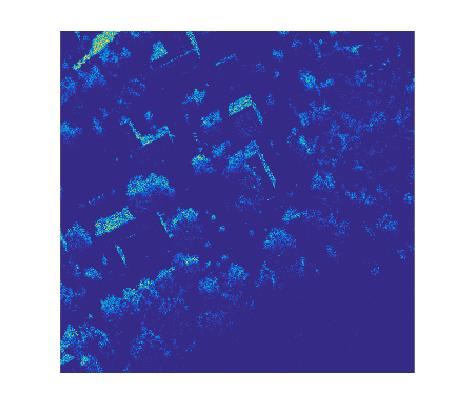}}
\subfigure[]{\includegraphics[height=2cm]{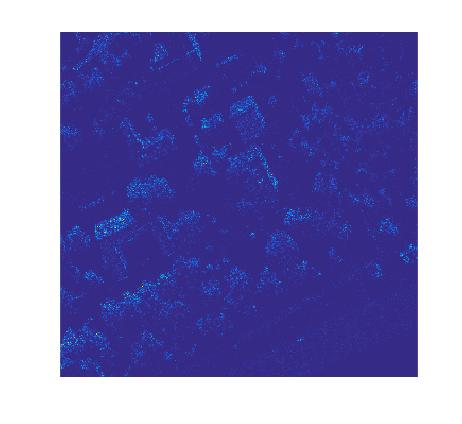}}
\subfigure[]{\includegraphics[height=2cm]{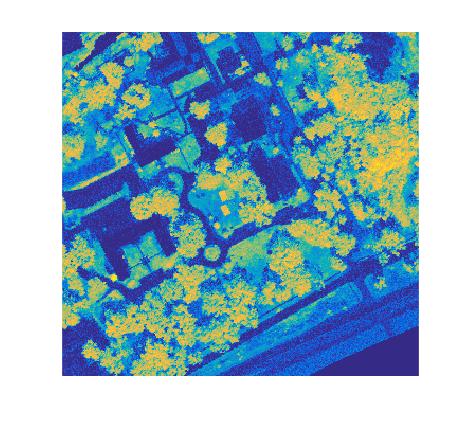}}

\caption{Estimated proportion maps using (a)-(g) supervised PM-LDA, (h)-(n) PM-LDA, (o)-(u) NCM-Bayes.}
\label{fig:p_gulfport}
\end{center}
\end{figure*}

 sPM-LDA was first applied to University of Pavia data set. This image was acquired in 2002 by Reflective Optics System Imaging Spectrometer (ROSIS) over Pavia, Italy. The scene contains $340 \times 610$ pixels, 103 bands with wavelengths ranging from 430 to 860 nm, and a spatial resolution of 1.3 m. An RGB image of the Pavia data set is shown in Fig \ref{fig:pavia}.

The Pavia HSI image was segmented into superpixels using the map-guided hyperspectral SLIC algorithm with the following parameter settings: $k = 500$ and $m = 20$. Fig \ref{fig:superpixels}(a) shows the superpixels generated by the hyperspectral SLIC approach. Then, building polygons selected and pulled from OpenStreetMap.org from the region of interest are shown in Fig \ref{fig:superpixels}(b). The final superpixels are shown in Fig \ref{fig:superpixels}(c) by merging all superpixels overlapping any building polygons.
 
The number of endmembers was set to 6 for the proposed method and comparison methods. As a pre-processing step, all the pixel signatures in the data set are normalized to have unit length as the input for all methods. Endmembers are initialized in sPM-LDA and PM-LDA using Vertex Component Analysis (VCA) \cite{nascimento2005vertex}. VCA was also used as the endmember extraction method applied in NCM-Bayes. Parameter settings for all methods were manually tuned  to yield the best performance. For sPM-LDA, $K = 6$, $\lambda = 1$, $\alpha = 0.3$, $\epsilon = 5\%$ and $T = 200$.  Blue roof and red roof materials were selected to be supervised in this study. To be more specific, the superpixels overlapping with the 3 blue roof or red roof polygons in Fig \ref{fig:superpixels}(b) were  labeled as such. For instance, superpixels in blue roof regions are given labels of $\tau = [1 0 1 1 1 1]$ while superpixels in red roof regions are given labels of $\tau = [0 1 1 1 1 1]$, where `1' in the first entry in $\tau$ denotes the possible existence of blue roof and `1' in the second entry denotes the possible existence of red roof, while `1' in the other four entries denotes the possible existence of the other four (unsupervised) endmembers. All other pixels in the scene are given $\tau = [0 0 1 1 1 1]$.  For NCM-Bayes, the Markov chain length was set to 250, the length of the burn-in period was set to 1000, $\delta = 0.001$, and the initial endmember variance was set to $0.001$.  For PM-LDA,  $K = 6$, $\lambda = 1$, $\alpha = 0.3$ and $T = 200$.

%\begin{figure*}
%\begin{center}
%\subfigure[]{\includegraphics[height=4cm]{E.jpg}}
%\subfigure[]{\includegraphics[height=4cm]{E_pmlda.jpg}}
%\subfigure[]{\includegraphics[height=4cm]{E_ncmbayes.jpg}}
%\caption{Endmember means estimated by (a) supervised PM-LDA, (b) PM-LDA, (c) VCA.}
%\label{fig:e_pavia}
%\end{center}
%\end{figure*}

For experimental results, all six endmember distribution parameters including endmember means and covariance matrices are estimated by sPM-LDA. The estimated endmember means are shown in Fig \ref{fig:p_pavia}. From these estimated endmember signatures and associated proportion maps, we can see that the signatures estimated by sPM-LDA are qualitatively more accurate on the Pavia data set.  For example, the blue roof endmember clearly aligns only with blue roof pixels in the scene.  The estimated proportion maps in Fig \ref{fig:p_pavia} illustrate the performance of the different methods. All proportion maps are permuted and aligned vertically according to the same major materials on each map so that each vertical map could be the same and compared. The six desired endmembers are blue roof, red roof, soil, asphalt, shadow and grass. Proportion maps estimated by supervised PM-LDA in the first row of Fig \ref{fig:p_pavia} are found to be smooth and the estimated proportion values are high for corresponding pixels dominated by single desired material and low for other materials. In contrast, (unsupervised) PM-LDA is  unable to separate blue roof from some other buildings and roads consisting of different materials across the scene. 

For quantitative evaluation, two evaluation metrics were computed: (1) entropy of proportion maps;  and (2) data likelihood given estimation results. The proportion entropy is defined in \eqref{eqn:entropy}.

\begin{equation} \label{eqn:entropy}
H\left(\mathbf{P}\right) = -\sum_{n=1}^{N} \sum_{k=1}^{K} p_{nk} \ln p_{nk} 
\end{equation}

where $p_{nk}$ is the proportion value for the $n^{th}$ pixel and $k^{th}$ endmember, $N$ is the number of pixels and $K$ is the number of endmembers. 
The motivation for applying proportion entropy is that in a hyperspectral image, there are usually a small number of  endmembers that are present in each pixel. Therefore, accurate proportion values for each pixels yield low proportion entropy indicating that only a few endmembers are present in each pixel. 
The second evaluation metric is that NCM log-likelihood over all pixels in the dataset. The NCM-likelihood provides a measure of the overall fit between the hyperspectral data points and endmember distriubtions under the NCM model. The metric is indicated in \eqref{eqn:likelihood}.

\begin{small}
\begin{equation} \label{eqn:likelihood}
%f\left(\mathbf{X}|\mathbf{E},\mathbf{P},\mathbf{\Sigma}\right) = \sum_{n=1}^{N} \frac{1}{\sqrt{(2\pi)^L |\sum_{k=1}^{K} p_{nk}^{2} \mathbf{\Sigma}_k |}} \text{exp}\left(-\frac{1}{2}(x_n-\sum_{k=1}^{K}p_{nk}\mathbf{\mu}_k)\right)
f\left(\mathbf{X}|\mathbf{E},\mathbf{P},\mathbf{\Sigma}\right) = \sum_{n=1}^{N} \ln \mathscr{N}\left(\mathbf{x}_n \Bigg|\sum_{k=1}^{K}p_{nk}\mathbf{e}_k,\sum_{k=1}^{K}p_{nk}^{2}\mathbf{\Sigma}_k\right)
\end{equation}
\end{small}

The quantitative evaluation using two metrics are shown in Table \ref{tab:entropy} and Tabel \ref{tab:likelihood}, respectively. 

\begin{table}[!htb]
\centering  \
\begin{tabular}{lccc} 
\hline
Dataset &NCM-Bayes &PM-LDA &sPM-LDA\\ \hline        
Pavia &2.23e5 &8.81e4 &\textbf{8.39e4}\\  \hline
\end{tabular}
\caption{Overall proportion map entropy for three methods on Pavia}
\label{tab:entropy}
\end{table}

\begin{table}[!htb]
\centering  \
\begin{tabular}{lccc} 
\hline
Dataset &NCM-Bayes &PM-LDA &sPM-LDA\\ \hline        
Pavia &8.97e6 &\textbf{6.68e7} &6.53e7\\  \hline
\end{tabular}
\caption{Overall log-likelihood for three methods on Pavia}
\label{tab:likelihood}
\end{table}

\subsubsection{Imprecise vs Precise Endmember Labels}

In the sPM-LDA results above, relatively precise labels of $\tau = [1 0 1 1 1 1]$ are given to indicate no red roof endmember exists for the superpixels in blue roof region by setting the second entry of $\tau$ to '0'. Similarly, relatively precise labels for superpixels in red roof regions were applied in the previous experiment. In this experiment,  these relatively ``precise'' labels were compared to relatively imprecise labels. For imprecise labels, the superpixels in blue roof and red roof regions are all given the same label vector, $\tau = [1 1 1 1 1 1]$, which relaxes the restriction for candidate endmembers in blue and red roof regions. All other pixels in non-blue-roof, non-red-roof regions are still labeled as $\tau = [0 0 1 1 1 1]$. In other words, blue and red roof materials were not distinguished from each other in the label vectors.  All other parameters are kept the same as in the original experiment. 

 Overall, the estimated endmember means and proportion maps by sPM-LDA with imprecise labels were found to have slightly more error when compared with those of sPM-LDA with precise labels. When examining the proportion maps in Fig \ref{fig:p_pavia_labels},  even with imprecise labels, the two different blue roof and red roof pixels are effectively separated. However, sidewalk pixels around the blue roof building are incorrectly partially represented by blue roof and red roof proportion values. This is the result of the fact that the sidewalk pixels are better represented by a linear combination of the blue and red roof endmembers than other endmembers (\emph{e.g.,} the asphalt signature). Therefore, as can be seen, sPM-LDA with imprecise labels can obtain good results but, of course, more precise labels can improve performance.  As expected, the quantitative results, both the entropy and likelihood values in Table \ref{tab:imprecise_pavia}, show that precise labels yield better unmixing performance than imprecise labels.

\begin{table}[!htb]
%\footnotesize
\centering  \
\begin{tabular}{lccc} 
\hline
Labels &Precise &Imprecise\\ \hline        
Entropy &\textbf{8.39e4} &8.81e4\\  \hline
Likelihood &\textbf{6.53e7} &6.46e7\\  \hline
\end{tabular}
\caption{Overall proportion map entropy and log-likelihood for sPM-LDA with precise and imprecise labels on Pavia.}
\label{tab:imprecise_pavia}
\end{table}

%\begin{figure}%[ht]
%\begin{center}
%\subfigure[]{\includegraphics[height=3cm]{E.jpg}}
%\subfigure[]{\includegraphics[height=3cm]{E_imprecise_pavia.jpg}}
%\caption[Estimated endmember means on Pavia with or without precise labels]{Endmember means estimated on Pavia by (a) semi-supervised PM-LDA with precise labels, (b) semi-supervised PM-LDA with imprecise labels.}
%\label{fig:e_pavia_labels}
%\end{center}
%\end{figure}

\begin{figure}[!h]
\begin{center}
\subfigure[]{\includegraphics[height=4cm]{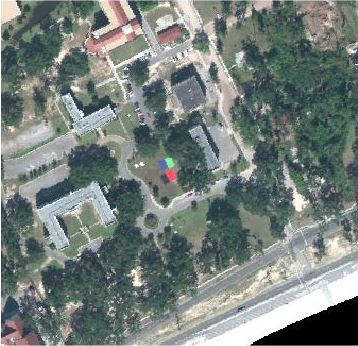}}
\subfigure[]{\includegraphics[height=4cm]{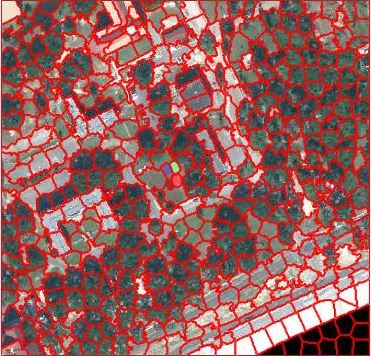}}
\caption{MUUFL Gulfport dataset: (a) RGB image; (b) superpixels by hyperspectral SLIC.}
\label{fig:rgb_gp}
\end{center}
\end{figure}

\subsection{\textit{MUUFL Gulfport}}

sPM-LDA was also applied to the MUUFL Gulfport data set \cite{Gulfport}. This image is acquired by the CASI-1500 hyperspectral imager flying over the University of Southern Mississippi-Gulfport in Long Beach, Mississippi in November 2010. The scene contains $325 \times 337$ pixels with 72 bands, of which wavelengths range from 375 to 1050 nm. The spatial resolution is 1 m. The RGB image of MUUFL Gulfport data set is shown in Fig \ref{fig:rgb_gp} (a). This dataset can be found on our Github page \cite{gulfportdata}.

The MUUFL Gulfport data were segmented into superpixels as is shown in Fig \ref{fig:p_pavia} (h). The parameters for superpixel segmentation was set to $k = 500$ and $m = 20$. The number of endmembers was set as seven for the proposed and comparison methods.  The parameter settings for all methods are selected to yield the best performance. For sPM-LDA, endmembers for red roof, light grey roof and beach sand were selected to be the semi-supervised endmembers in this study. Parameters used in sPM-LDA were set to be: $K = 7$, $\lambda = 1$, $\alpha = 0.3$, $\epsilon = 10\%$ and $T = 200$. For NCM-Bayes, the Markov chain length was set to 250, the length of the burn-in period was set to 1000, $\delta = 0.001$, and the initial endmember variance was set to $0.001$.  For PM-LDA,  $K = 7$, $\lambda = 1$, $\alpha = 0.3$ and $T = 200$.

Similar to experiments on Pavia, seven endmember distribution parameters including endmember means and covariance matrices were estimated. The seven  endmembers corresponded to red roof, soil, light grey roof, shadow, asphalt, beach sand and vegetation materials. The estimated endmember means are shown in Fig \ref{fig:e_gulfport} and proportion maps are shown in Fig \ref{fig:p_gulfport}. All desired materials are found in proportion maps of sPM-LDA. For PM-LDA, the red roof and light grey roof are not accruately estimated. For NCM-Bayes, soil is mixed with most materials in the dataset as is shown in (p) and shadow is mapped to three endmembers from (r) to (t).

\begin{table}[!htb]
\centering  \
\begin{tabular}{lccc} 
\hline
Dataset &NCM-Bayes &PM-LDA &sPM-LDA\\ \hline        
Pavia &1.15e5 &4.18e4 &\textbf{3.96e4}\\  \hline
\end{tabular}
\caption{Overall proportion map entropy for three methods on Gulfport.}
\label{tab:entropy_gp}
\end{table}

\begin{table}[!htb]
\centering  \
\begin{tabular}{lccc} 
\hline
Dataset &NCM-Bayes &PM-LDA &sPM-LDA\\ \hline        
Pavia &3.28e6 &\textbf{2.09e7} &1.97e7\\  \hline
\end{tabular}
\caption{Overall log-likelihood for three methods on Gulfport}
\label{tab:likelihood_gp}
\end{table}

\section{Summary}
\label{sec:conclusion}
In this paper, a semi-supervised hyperspectral unmixing method modeled by PM-LDA is presented. This approach (sPM-LDA) is capable of incorporating imprecise labels in a semi-supervised hyperspectral unmixing process to obtain more sparse and interpretable endmember distribution and proportion estimations. 
%\section{REFERENCES}
%\label{sec:refs}

\bibliographystyle{IEEEbib}
\bibliography{strings,refs}

\end{document}